\pdfoutput=1
\documentclass[letterpaper]{article} 
\usepackage{aaai23}  
\usepackage{times}  
\usepackage{helvet}  
\usepackage{courier}  
\usepackage[hyphens]{url}  
\usepackage{graphicx} 
\usepackage{natbib}  
\usepackage{caption} 
\usepackage{algorithm}
\usepackage{algorithmic}

\usepackage{amsmath}
\usepackage{amssymb}
\usepackage{multirow}
\usepackage{mathrsfs}
\usepackage{booktabs}

\usepackage{newfloat}
\usepackage{listings}
\DeclareCaptionStyle{ruled}{labelfont=normalfont,labelsep=colon,strut=off} 
\floatstyle{ruled}
\newfloat{listing}{tb}{lst}{}
\floatname{listing}{Listing}
\title{Painterly Image Harmonization in Dual Domains}
\author{
    Junyan Cao,
    Yan Hong,
    Li Niu\thanks{Corresponding author.}\\
}
\affiliations{
    MoE Key Lab of Artificial Intelligence, Shanghai Jiao Tong University\\

    \{joy\_c1, ustcnewly\}@sjtu.edu.cn, yanhong.sjtu@gmail.com
}

\usepackage{bibentry}

\begin{document}

\maketitle

\begin{abstract}
Image harmonization aims to produce visually harmonious composite images by adjusting the foreground appearance to be compatible with the background. When the composite image has photographic foreground and painterly background, the task is called painterly image harmonization. There are only few works on this task, which are either time-consuming or weak in generating well-harmonized results. 
In this work, we propose a novel painterly harmonization network consisting of a dual-domain generator and a dual-domain discriminator, which harmonizes the composite image in both spatial domain and frequency domain. The dual-domain generator performs harmonization by using AdaIN modules in the spatial domain and our proposed ResFFT modules in the frequency domain. The dual-domain discriminator attempts to distinguish the inharmonious patches based on the spatial feature and frequency feature of each patch, which can enhance the ability of generator in an adversarial manner. 
Extensive experiments on the benchmark dataset show the effectiveness of our method. Our code and model are available at \url{https://github.com/bcmi/PHDNet-Painterly-Image-Harmonization}.
\end{abstract}

\section{Introduction}\label{sec:intro}

\begin{figure}[tp!]
\centering
\includegraphics[width=0.99\linewidth]{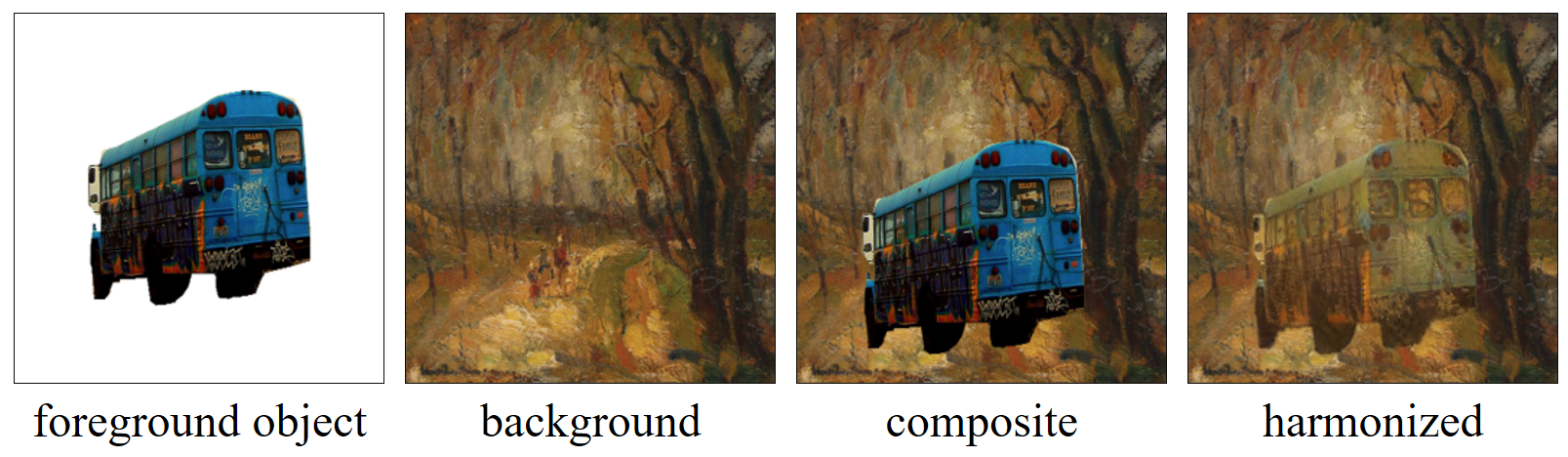}
\caption{Example of painterly image harmonization. From left to right are foreground object, background painting image, composite image, and harmonized image. }
\label{fig:examples}
\end{figure}

Image composition refers to cutting the foreground from one image and pasting it on another background image, producing a composite image.
However, the foreground and background may have inconsistent color and illumination statistic, making the whole composite image inharmonious and unrealistic. Image harmonization \cite{lalonde2007using,tsai2017deep,cong2020dovenet} aims to adjust the foreground appearance to make it compatible with the background.
In recent years, image harmonization has attracted growing research interest \cite{bargain,guo2021image,hang2022scs}. Besides combining foreground and background from photos, users may insert an object into a painting for creative painterly editing. This task is called painterly image harmonization, which has only received limited attention \cite{luan2018deep,zhang2020deep,peng2019element}. In particular, when a composite image is composed of photographic foreground object and  painterly background image, painterly image harmonization aims to adjust the foreground style in the composite image to produce a harmonious image. 
Figure~\ref{fig:examples} shows an example of painterly image harmonization.

Existing painterly image harmonization methods can be divided into optimization-based methods \cite{luan2018deep, zhang2020deep} and feed-forward methods \cite{peng2019element}. Optimization-based methods directly optimize the composite image to minimize the designed objective function. Specifically, the optimization-based methods \cite{luan2018deep, zhang2020deep} employ a set of losses (\emph{e.g.}, content loss \cite{gatys2016image}, style loss \cite{huang2017arbitrary}, smoothness loss \cite{mahendran2015understanding}) as the objective function. Then for each input composite image, they iteratively update its pixels and output the harmonized result, which does not rely on any training data. However, the optimization-based methods \cite{peng2019element} are very time-consuming, which could not achieve real-time harmonization. Feed-forward methods pass the composite image through the network to produce a harmonized image. 
In particular, they train the network on the training set with a set of losses. However, the foregrounds are often not sufficiently stylized or not naturally blended into the background. Considering the demand of real-time application, we follow the research line of feed-forward method, which is also dominant in artistic style transfer \cite{huang2017arbitrary,park2019arbitrary,liu2021adaattn}.



In this work, we perform painterly image harmonization in two domains: spatial domain and frequency domain. Unlike previous works which only consider spatial domain \cite{luan2018deep,zhang2020deep,peng2019element}, we additionally explore frequency domain due to the following two concerns. Firstly, the convolution operations in spatial domain have local receptive field, and lack the ability to capture long-range dependency \cite{wang2018non}. Meanwhile, the operations in frequency domain, \emph{e.g.}, Fast Fourier Transform (\emph{FFT}), have image-wise receptive field and thus could extract the global style of the whole image. Secondly, painterly image harmonization needs to transfer the style (\emph{e.g.}, color, stroke, pattern, texture) of background image to the composite foreground. 
The background paintings often have periodic textures and patterns which appear regularly, which could be well captured in the frequency domain.

Motivated by the advantage of frequency domain, we propose a novel dual-domain network named \textbf{PHD}Net to accomplish \textbf{P}ainterly image \textbf{H}armonization in \textbf{D}ual domains. Our PHDNet consists of a dual-domain generator and a dual-domain discriminator. Specifically, our generator is built upon UNet~\cite{ronneberger2015u}. We harmonize multi-scale encoder feature maps in the spatial domain and frequency domain sequentially in the skip connections. We first apply Adaptive Instance Normalization (AdaIN) \cite{huang2017arbitrary} to align the statistics (\emph{i.e.}, mean and variance) of composite foreground feature map with background feature map in the spatial domain. Then, we convert the normalized feature map to frequency feature map and apply our proposed ResFFT module to harmonize the frequency feature map in the frequency domain.  
For our dual-domain discriminator, we divide the composite image into different patches including foreground patches and background patches. We extract the spatial domain feature and frequency domain feature for each patch. Based on the dual-domain patch features, the discriminator strives to distinguish the foreground patches from background patches, while the generator attempts to fool the discriminator. The dual-domain discriminator can promote the harmonization ability of dual-domain generator in an adversarial manner, so that the foreground in the harmonized image is inseparable from the background and appears to exist in the original painting. We conduct extensive experiments to verify the effectiveness of our proposed dual-domain network.
Our contributions are summarized as follows,

\begin{itemize}
    \item 
    To the best of our knowledge, we are the first to introduce frequency domain knowledge into painterly harmonization task.
    
    \item We accomplish painterly image harmonization in dual domains, and design a dual-domain network PHDNet. Our PHDNet contains a dual-domain generator with a novel ResFFT module to harmonize the composite image in both spatial and frequency domain, and a novel dual-domain discriminator to distinguish the inharmonious region in both spatial and frequency domain.
    
    \item Comprehensive experiments demonstrate that our PHDNet could produce more harmonious results with consistent style and intact content than previous methods.
\end{itemize}

\section{Related Work}

\subsection{Image Harmonization}

Image harmonization aims to harmonize a composite image with both foreground and background from photos. 
Early traditional image harmonization methods \cite{song2020illumination,xue2012understanding,multi-scale,lalonde2007using}
focused on manipulating the low-level statistics  (\emph{e.g.}, color, gradient, histogram) of foreground to match those of background. Then, unsupervised deep learning methods \cite{zhu2015learning} adopted adversarial learning to enforce the harmonized images to be indistinguishable from real images. More recently, abundant supervised deep learning methods~\cite{tsai2017deep,sofiiuk2021foreground,Cong_2022_CVPR} leveraged paired training data to train harmonization models. To name a few, \citet{xiaodong2019improving} and \citet{Hao2020bmcv} designed various attention mechanisms. \citet{cong2020dovenet} and \citet{bargain} formulated image harmonization as domain translation task by treating foreground and background as two domains. \citet{guo2021intrinsic} and \citet{guo2021image} decomposed an image to reflectance map and illumination map, followed by adjusting the foreground illumination map.
\citet{ling2021region} and \citet{hang2022scs} introduced AdaIN \cite{huang2017arbitrary} in style transfer to image harmonization. 
The above supervised image harmonization methods require ground-truth images as supervision, which are not applicable to our task.

\subsection{Painterly Image Harmonization}
When inserting an object into a painting, painterly image harmonization aims to transfer the background style to the foreground while retaining the foreground content, making the composite image as natural as possible. As far as we are concerned, there are only few works on painterly image harmonization.
\citet{luan2018deep} proposed to migrate relevant statistics
of neural responses to the inserted object, ensuring both spatial and inter-scale statistical consistency.
\citet{zhang2020deep} developed a novel Poisson gradient loss jointly optimized with content and style loss.
\citet{peng2019element} employed AdaIN to manipulate the foreground feature map, together with global and local discriminators for adversarial learning.
All these methods only considered spatial domain, while we perform harmonization in both spatial domain and frequency domain. 

\subsection{Artistic Style Transfer}
The goal of artistic style transfer is stylizing a content image according to the provided style image. The existing style transfer methods can also be divided into optimization-based methods and feed-forward methods. The optimization-based methods \cite{gatys2016image,li2017demystifying,kolkin2019style,du2020much} proposed to optimize over the content image to match its style with style image. The feed-forward methods combined the content of content image and the style of style image to produce a stylized image, during which the style-relevant statistics (\emph{e.g.}, mean, variance) between foreground features and background features are matched in the network. According to global matching and local matching (matching corresponding regions), the feed-forward methods can be further divided into global transformation  methods \cite{huang2017arbitrary,li2017universal,li2018learning} and local transformation methods \cite{park2019arbitrary,liu2021adaattn,huo2021manifold,deng2022stytr2}.
Different from the above methods which stylize the entire content image, painterly image harmonization needs to consider the location of inserted object and harmonize it accordingly, achieving the goal that the object appears to be present in the original painting.

\subsection{Frequency Domain Learning}
Frequency domain information has been exploited in deep learning based methods for myriads of computer vision tasks, due to its enticing properties (\emph{e.g.}, large receptive field, high and low frequency separation). For instance, a few works \cite{xu2020learning,roy2021image,shen2021dct} converted the input image or output mask of network to frequency domain. Similarly,  \citet{suvorov2022resolution} and \citet{mardani2020neural} converted the intermediate features in the network to frequency domain, and processed the frequency features to achieve the goals of different tasks. By decomposing an image to low-frequency part and high-frequency part, some recent works \cite{bansal2017pixelnn,yang2020fda,yu2021wavefill,cai2021frequency} proposed to manipulate the structural information and detailed information separately. In this work, we make the first attempt to introduce frequency domain into painterly image harmonization. 

\begin{figure*}[tp!]
\centering
\includegraphics[width=0.96\linewidth]{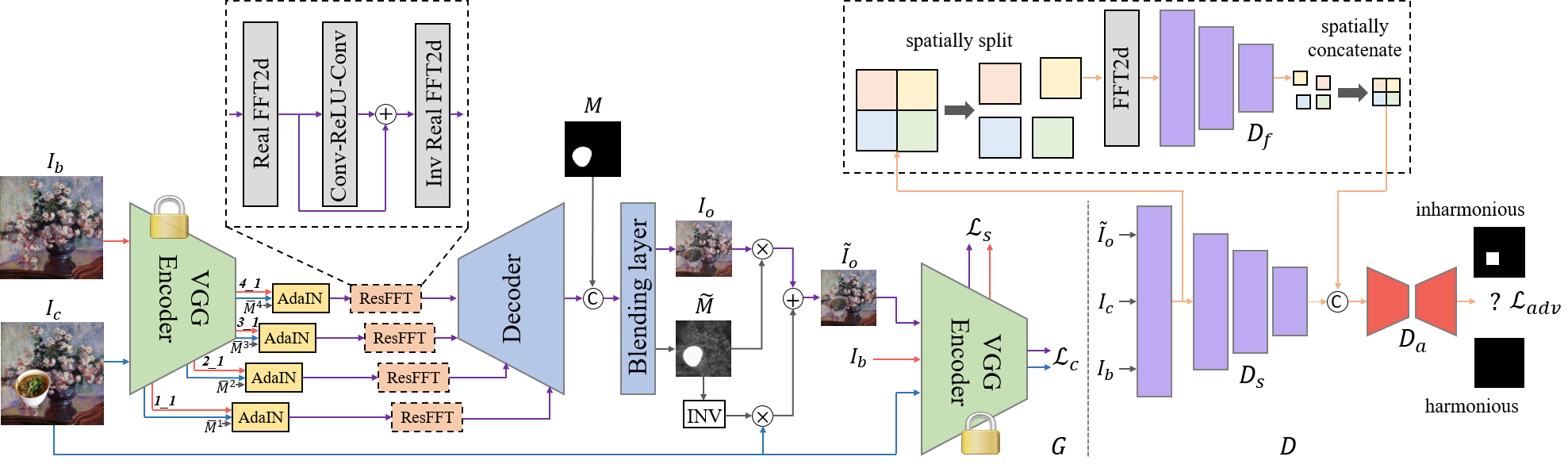}
\caption{The architecture of our PHDNet, which consists of a dual-domain generator $G$ and a dual-domain discriminator $D$. Pretrained VGG encoder is freezed. ``INV'' means ``inverse''.}
\label{fig:network}
\end{figure*}


\section{Our Method}

The architecture of our PHDNet is shown in Figure~\ref{fig:network}. A composite image ${I}_{c}$ is obtained by pasting foreground object ${I}_{c}^{f}$ on a complete background painting ${I}_{b}$, and we use a foreground mask $M$ to indicate the foreground region. Our goal is to train a model that transfers the style from ${I}_{b}$ to ${I}_{c}^{f}$ while keeping the content of ${I}_{c}^{f}$, generating a harmonized image $\tilde {I}_{o}$.

Our PHDNet consists of a dual-domain generator and a dual-domain discriminator, under the adversarial learning framework~\cite{goodfellow2014generative}. As demonstrated in Figure~\ref{fig:network}, the dual-domain generator $G$ takes in ${I}_{c}$ and ${I}_{b}$, and adjusts the style of ${I}_{c}^{f}$ in both spatial domain and frequency domain. 
We also employ a dual-domain discriminator $D$, 
which predicts an inharmonious mask to distinguish the foreground patches from the background patches. The discriminator $D$ is used to strengthen the generator $G$ in an adversarial manner. Next, we will detail our dual-domain generator in Section~\ref{sec:generator} and dual-domain discriminator in Section~\ref{sec:discriminator}.

\subsection{Dual-Domain Generator} \label{sec:generator}


We employ the encoder-decoder architecture in \cite{huang2017arbitrary} as our backbone, in which the encoder is pretrained VGG-19 network~\cite{VGG19} and the decoder is a symmetric structure of encoder. Note that we only use the first few layers (up to \emph{ReLU-4\_1}) of VGG-19 as our encoder, and freeze them to extract multi-scale encoder features. 
Following \cite{ronneberger2015u}, we add skip connections on \emph{ReLU-1\_1}, \emph{ReLU-2\_1}, and \emph{ReLU-3\_1} layers of the encoder. 
By feeding ${I}_{c}$ and ${I}_{b}$ into the encoder respectively, we could get the feature map ${F}_{gc}^{l}$ and ${F}_{gb}^{l}$ extracted by the $l$-th layer (\emph{l} $\in \left\{1, 2, 3, 4\right\}$) of encoder. The four encoder layers contain \emph{ReLU-1\_1}, \emph{ReLU-2\_1}, and \emph{ReLU-3\_1}, and \emph{ReLU-4\_1} (bottleneck). Then for the $l$-th layer, we feed ${F}_{gc}^{l}$ and ${F}_{gb}^{l}$ jointly with a downsampled mask $\bar{M}^{l}$ into the AdaIN module followed by our ResFFT module, aiming to transfer the style from ${I}_{b}$ to ${I}_{c}^{f}$ in both spatial domain and frequency domain. 
Detailed architectures of these two modules will be introduced later. The harmonized encoder features are taken as the input of decoder or concatenated with decoder features via skip connection. 
At the end of decoder, we insert a blending layer similar to \cite{sofiiuk2021foreground}, which takes the concatenation of the decoder feature map and mask $M$ as input, producing a soft mask $\tilde{M}$ for the final blending.

\subsubsection{AdaIN Module} \label{adain} 

Firstly, we apply AdaIN~\cite{huang2017arbitrary} in the spatial domain. As stated above, the input of AdaIN module contains three parts: the foreground mask, the encoder feature maps of both composite image and background image. 

Inspired by \cite{huang2017arbitrary,ling2021region}, for the $l$-th layer of VGG-19 encoder, we pass ${F}_{gc}^{l}$ and ${F}_{gb}^{l}$ jointly with $\bar {M}^{l}$ through the AdaIN module in Figure~\ref{fig:network},
aiming to align the channel-wise mean and standard deviation of the foreground region of ${F}_{gc}^{l}$ to those of the whole region of ${F}_{gb}^{l}$. The process could be expressed as
\begin{equation}\label{eq:norm} 
    \begin{aligned}
    {F}_{gs}^{l} =\left(\sigma(F_{gb}^{l})\frac{{F}_{gc}^{l}-\mu(F_{gc}^{l} \circ \bar M^l)}{\sigma(F_{gc}^{l} \circ \bar M^l)}+\mu(F_{gb}^{l})\right) \circ \bar M^l \\
    + {F}_{gc}^{l} \circ \left(1 - \bar M^l\right),
    \end{aligned}
\end{equation}
where $\circ$ means element-wise multiplication, $\mu(\cdot)$ and $\sigma(\cdot)$ denote the formulas to compute the mean and standard deviation of the feature map within the masked region (see \cite{huang2017arbitrary,ling2021region} for details). 

\subsubsection{ResFFT Module} \label{res_fft}
Then we feed the normalized feature map ${F}_{gs}^{l}$ into our ResFFT module for harmonization in the frequency domain.
Following \cite{suvorov2022resolution}, we apply \emph{Real FFT} to feature map $F_{gs}^l$ with size $h^l\times w^l \times c_g^l$ and drop the redundant negative frequency terms due to the symmetric property, leading to the frequency feature map. 
The obtained frequency feature map is in the complex form with two parts, \emph{i.e.}, real and imaginary part, both of which have the size $h^l \times \frac{w^l}{2} \times c_g^l$. We concatenate two parts channel-wisely and obtain the frequency feature map $F_{gf}^l$ with size $h^l \times \frac{w^l}{2} \times 2c_g^l$.

Then we pass frequency feature map $F_{gf}^l$ through the residual block \cite{he2016deep}. In the residual block, we learn the residual and add it to the input frequency feature map. Intuitively, we hope that the learned residual could harmonize the frequency feature map, \emph{e.g.}, restoring the missing or corrupted texture and pattern within the foreground region in the frequency domain. Through the residual block, we get the harmonized frequency feature map  $\hat{F}_{gf}^l$.
Finally we convert $\hat{F}_{gf}^l$ back to the spatial domain. In detail, we first convert $\hat{F}_{gf}^l$ to complex form and
then apply \emph{inverse Real FFT} to get the spatial feature map $\hat{F}_{gs}^l$ with size $h^l \times w^l \times c_g^l$.

After AdaIN module and ResFFT module, we obtain the harmonized feature map $\hat{F}_{gs}^l$, which is delivered to the decoder to generate the coarse output $I_o$. Then we blend $I_o$ with the composite image $I_c$ using the soft mask $\tilde M$, producing the harmonized image $\tilde I_o$:
\begin{equation}\label{eq:gen_output}
    \begin{aligned}
    \tilde{I}_{o}={I}_{o} \circ \tilde{M} + {I}_{c} \circ (1-\tilde{M}),
    \end{aligned}
\end{equation}
where $\tilde M$ is produced by the blending layer as mentioned above.

To match multi-scale style statistics between the background image and the foreground of harmonized image, we employ the style loss in~\cite{huang2017arbitrary}, which could be expressed as
\begin{equation}\label{eq:loss_s}
\begin{aligned}
    \mathcal{L}_{s}=\sum_{l=1}^{L}\left\|\mu\left(\phi^{l}(\tilde{I}_{o})\circ \bar M^l \right) -\mu\left(\phi^{l}({I}_{b})\right)\right\|^{2} + \\
\sum_{l=1}^{L}\left\|\sigma\left(\phi^{l}(\tilde{I}_{o})\circ \bar M^l \right)-\sigma\left(\phi^{l}({I}_{b})\right)\right\|^{2},
\end{aligned}
\end{equation}
where each $\phi^{l}, l\in\left\{1,2,3,4\right\}$ denotes the $l$-th \emph{ReLU-l\_1} layer in VGG-19 encoder.

Besides, we utilize a content loss \cite{gatys2016image} to ensure that the content of $\tilde I_o$ is close to that of $I_c$:
\begin{equation}\label{eq:loss_c}
    \begin{aligned}
    \mathcal{L}_{c}=\left\|\phi^4(\tilde{I}_{o})-\phi^4({I}_{c})\right\|^2.
    \end{aligned}
\end{equation}


\subsection{Dual-Domain Discriminator} \label{sec:discriminator}

To improve the quality of harmonized image $\tilde{I}_{o}$, we resort to adversarial learning and design a dual-domain discriminator $D$. Given an input image uniformly split into $n\times n$ patches, $D$ contains an encoder with spatial (\emph{resp.}, frequency) branch $D_s$ (\emph{resp.}, $D_f$) to extract the spatial (\emph{resp.}, frequency) feature for each patch, followed by a light-weighted auto-encoder $D_a$ to identify the inharmonious patch. Detailed architectures could be found in the Supplementary.

As shown in Figure~\ref{fig:network}, given an input image $I$, we pass it through the spatial branch $D_s$ to get the bottleneck feature map $F_{ds}$ with size $n\times n\times c_{ds}$, in which each pixel-wise feature vector in $F_{ds}$ is deemed as the spatial feature vector for one patch. 

Then we choose one intermediate feature map $F_{dm}$ in $D_s$ and derive the frequency feature for each patch. 
Supposing that $F_{dm}$ has size $m \times m \times c_{dm}$, we uniformly divide $F_{dm}$ into $n\times n$ non-overlapped patches with patch size being $(\frac m n) \times (\frac m n) \times c_{dm}$. We denote the $(i,j)$-th patch in $F_{dm}$ as $F_{dm}^{i,j}$, in which $i,j \in [1,n]$. Similar to the ResFFT module in Section \ref{sec:generator}, we apply \emph{FFT} to each patch separately and convert it to frequency domain. In particular, for $F_{dm}^{i,j}$, we obtain the converted frequency feature map $F_{df}^{i,j}$ containing the real part and the imaginary part, both of which have the size $(\frac m n) \times (\frac m n) \times c_{dm}$. Note that we use both positive and negative frequency terms here for regular feature map size. Then we concatenate the real and imaginary parts of $F_{df}^{i,j}$ channel-wisely and feed it into $D_f$ to get a $c_{df}$-dim frequency feature vector $\hat f_{df}^{i,j}$. Each frequency feature vector $\hat f_{df}^{i,j}$ encodes the frequency domain information of the $(i,j)$-th patch. 

We spatially combine $\hat f_{df}^{i,j}$ according to the spatial position $(i,j)$, yielding a frequency feature map $\hat{F}_{df}$ with size $n\times n\times c_{df}$. We concatenate $\hat{F}_{df}$ with $F_{ds}$ to form a feature map with size $n\times n\times (c_{df}+c_{ds})$, in which each pixel-wise feature vector contains both spatial domain information and frequency domain information of one patch. 
Then, the concatenated feature map is delivered to $D_a$ to predict a $n\times n$ inharmonious region mask, in which $0$ indicates harmonious patches and $1$ indicates inharmonious patches.  

By taking the harmonized result $\tilde I_o$, the composite image $I_c$, and the background image $I_b$ as the input of $D$ separately, we could get a $n\times n$ inharmonious region mask for each input. The loss function to update $D$ could be written as
\begin{equation}\label{eq:loss_d} \small
    \begin{aligned}
\mathcal{L}^{D}_{adv} = \|D(\tilde{I}_{o})-\bar M\|^2+\|D({I}_{c})-\bar M\|^2+
\|D({I}_{b})\|^2,
    \end{aligned}
\end{equation}
where $\bar M$ means the downsampled mask with size $n \times n$. For $I_c$ and $\tilde{I}_{o}$, we expect that $D_a$ could distinguish the foreground (inharmonious) patches from the background (harmonious) patches, so the predicted inharmonious region mask aligns with $\bar M$. For ${I}_b$, since there is no inharmonious patch, the predicted inharmonious region mask is supposed to be an all-zero mask. 

Under the adversarial learning framework \cite{goodfellow2014generative}, we update the dual-domain generator $G$ and the dual-domain discriminator $D$ alternatingly. When updating $G$, we hope that the harmonized output $\tilde I_o$ could confuse $D$, so that $D$ is unable to distinguish the inharmonious patches. Therefore, the adversarial loss to update $G$ is given as $\mathcal{L}^{G}_{adv}=\|D(\tilde{I}_{o})\|^2$. 

So far, the total loss for training $G$ is summarized as 
\begin{equation}\label{eq:loss_g}
    \begin{aligned}
\mathcal{L}_{G} =\mathcal{L}_{s}+\lambda_{c}\mathcal{L}_{c}+\lambda_{adv}\mathcal{L}_{G_{adv}},
    \end{aligned}
\end{equation}
where the trade-off parameters $\lambda_{c}$ and $\lambda_{adv}$ are set to $2$ and $10$ respectively in our experiments.

\begin{figure*}[t!]
\centering
\includegraphics[width=0.99\linewidth]{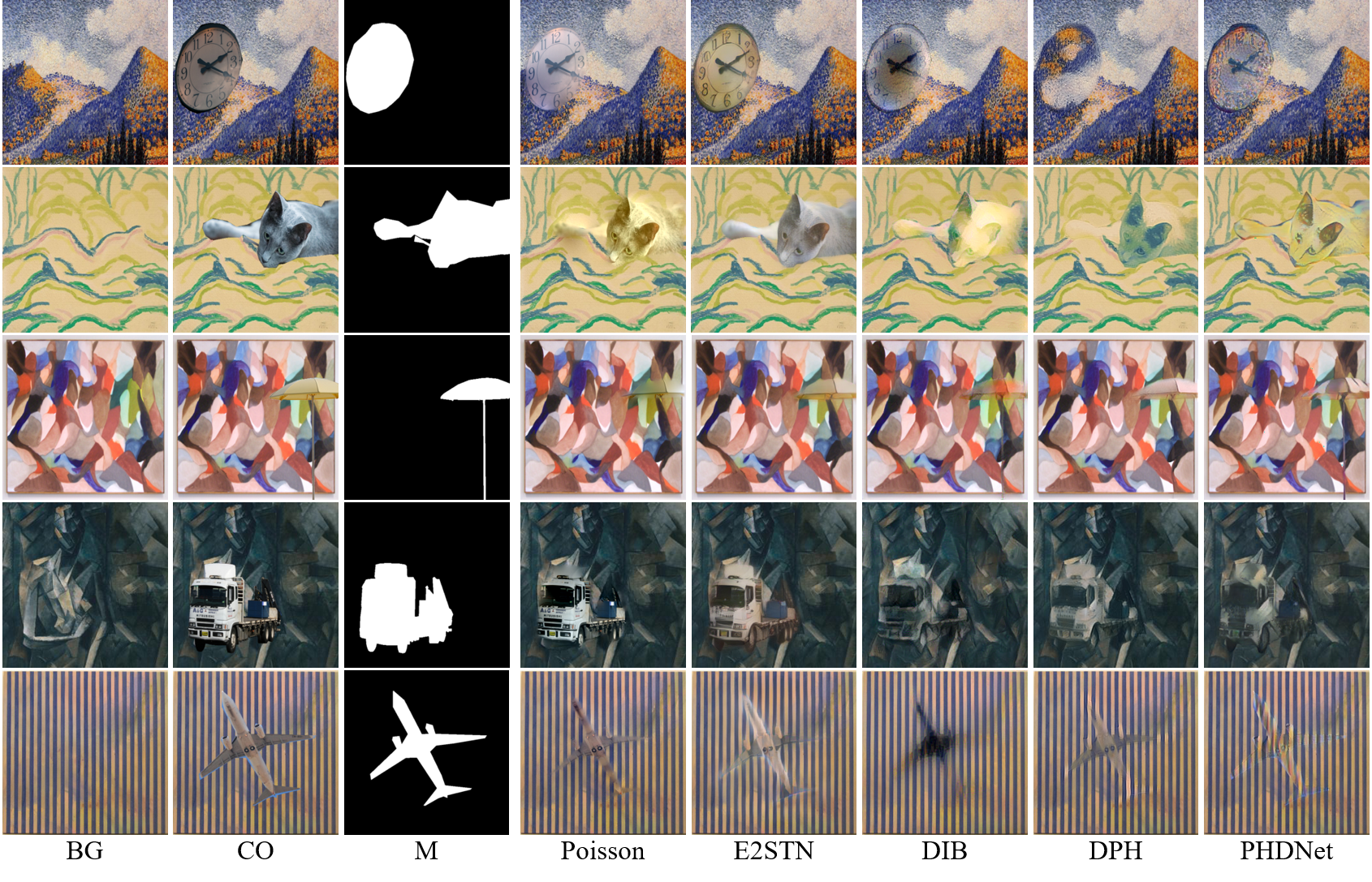}
\caption{Example results of painterly image harmonization baselines and our PHDNet. ``BG'' (\emph{resp.}, ``CO'') means ``background'' (\emph{resp.}, ``composite''). 
}
\label{fig:baseline_painterly}
\end{figure*}

\begin{figure*}[t!]
\centering
\includegraphics[width=0.99\linewidth]{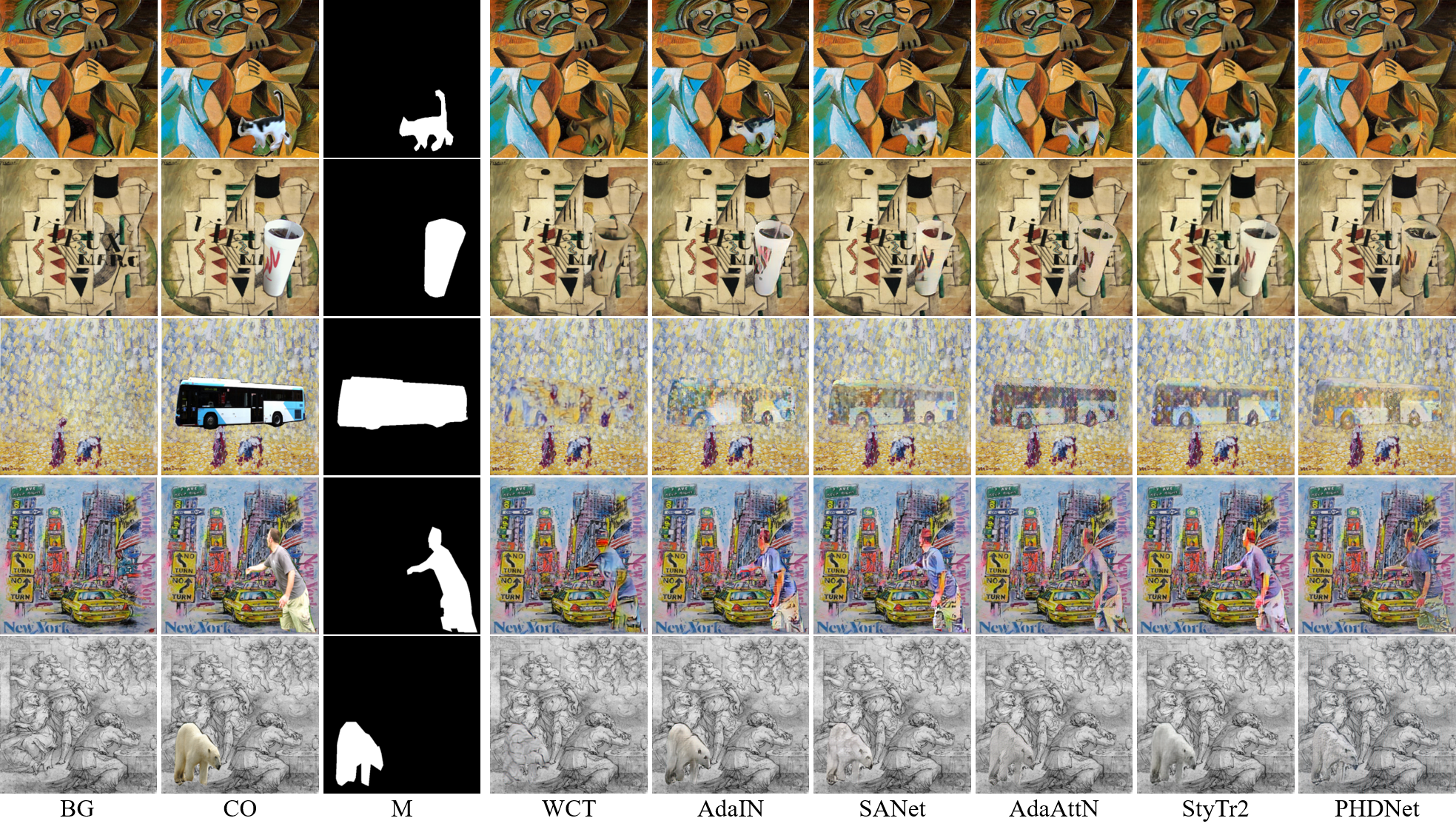}
\caption{Example results of artistic style transfer baselines and our PHDNet. 
}
\label{fig:baseline_style}
\end{figure*}

\section{Experiments}
We conduct experiments on COCO \cite{lin2014microsoft} and WikiArt \cite{artgan2018}. Refer to the Supplementary for more implementation details.


\subsection{Baselines}
We divide baselines into two groups: painterly image harmonization methods~\cite{luan2018deep,zhang2020deep,peng2019element} and artistic style transfer methods~\cite{huang2017arbitrary,liu2021adaattn}. 

The first group of baselines process the foreground region of  composite image. 
We compare with Deep Image Blending~\cite{zhang2020deep} (``DIB'' for short), Deep Painterly Harmonization~\cite{luan2018deep} (``DPH'' for short) and E2STN~\cite{peng2019element}. We also include traditional image blending method Poisson Image Editing~\cite{poisson} (``Poisson'' for short) for comparison.

The second group of baselines stylize the whole photographic (content) image which provides the foreground object. To adapt artistic style transfer methods to our task, we first stylize the entire content image according to the background (style) image. Then we cut the stylized foreground object and paste it on the background image.
We compare with typical or recent style transfer methods: WCT~\cite{li2017universal}, AdaIN~\cite{huang2017arbitrary}, SANet~\cite{park2019arbitrary}, AdaAttN~\cite{liu2021adaattn}, and StyTr2~\cite{deng2022stytr2}.

\subsection{Qualitative Analysis}
To compare with the first group of baselines, the results of different methods are illustrated in Figure~\ref{fig:baseline_painterly}. Although Poisson~\cite{poisson} could smoothen the boundary, the styles of foreground and background are still dramatically different. 
E2STN~\cite{peng2019element} is also struggling to transfer the style (\emph{e.g.}, row 2, 4).
DIB~\cite{zhang2020deep} could transfer the style to some extent, but it severely distorts the content information of foreground object (\emph{e.g.}, row 2, 5). 
DPH~\cite{luan2018deep} achieves competitive results among the baselines. 
Compared with DPH~\cite{luan2018deep}, our PHDNet can preserve the content structure better (row 1) and transfer the style better (row 2). Our PHDNet also has stronger ability to transfer the texture and pattern from background image. For example, our PHDNet can transfer the colorful strips to the umbrella (row 3) and quadrangle color blocks with sharp edges to the truck (row 4). Our PHDNet can also restore the vertical strips in the foreground region (row 5).

To compare with the second group of baselines, the results of different methods are illustrated in Figure~\ref{fig:baseline_style}. 
Since style transfer methods stylize the entire content image with limited attention paid to the foreground object, the foreground object may not be sufficiently stylized (row 1, 4),  which makes the foreground very obtrusive and easily separated from the background. In contrast, our PHDNet focuses on the foreground stylization. By taking the locality into account, in our harmonized results, the foreground object has more consistent style with its neighboring regions and thus appears to be more naturally blended into the background. Moreover, our PHDNet has stronger style transfer ability. For example, in row 1, the background has several green spots, so the foreground cat also has green spots. In row 2, 4, 5, the foreground objects of other methods are very smooth while our foreground objects own the fine-grained texture transferred from background images. 

The advantages of our PHDNet come from two aspects. Firstly, we perform harmonization in both spatial domain and frequency domain. As claimed in Section~\ref{sec:intro}, the frequency feature can capture the global style and periodic texture/pattern, so our PHDNet is able to reconstruct the missing or corrupted textures and patterns in the foreground. Secondly, the discriminator helps the generator in an adversarial manner, so that the foreground in the harmonized image is more compatible with the background. 

\subsection{User Study} \label{sec:user_study}
As there is no ground-truth harmonized image, we cannot use evaluation metrics (\emph{e.g.}, MSE, PSNR) to evaluate the model performance quantitatively. Therefore, we conduct user study to compare different methods.
We randomly select 100 content images from COCO and 100 style images from WikiArt to generate 100 composite images for user study. We compare the harmonized results generated by SANet~\cite{park2019arbitrary}, AdaAttN~\cite{liu2021adaattn}, StyTr2~\cite{deng2022stytr2}, DPH~\cite{luan2018deep}, E2STN~\cite{peng2019element}, and our PHDNet.

Specifically, for each composite image, we can obtain $6$ harmonized outputs generated by $6$ above-mentioned methods. Then we select $2$ images from these $6$ images to construct image pairs. Based on 100 composite images, we could construct 1,500 image pairs. Then we invite 20 users to see one image pair each time and pick out the more harmonious one. Finally, we collect 30,000 pairwise results and employ the Bradley-Terry (B-T) model~\cite{bradley1952rank,lai2016comparative} to obtain the overall ranking of all methods. The results are reported in Table~\ref{tab:bt_score} in the left sub-table, in which we can observe that our PHDNet achieves the highest B-T score and outperforms other baselines.

\begin{table}[tb] \small
\centering
\resizebox{\linewidth}{!}{
\begin{tabular}{c|c|c}
\toprule  Method & Type & B-T \\
\hline
DPH & OP & 0.555 \\
E2STN & FF & -1.811 \\
SANet & FF & -0.168 \\
AdaAttN & FF & 0.029 \\
StyTr2 & FF & 0.343 \\
PHDNet & FF & 1.052 \\
\bottomrule
\end{tabular}

\hspace{0.1em}

\begin{tabular}{c|c|c|c}
\toprule 
\multirow{2}*{\#} & \multicolumn{1}{c|}{$G$} & \multicolumn{1}{c|}{$D$} & \multirow{2}*{B-T}  \\
\cline{2-3}
~ &  w/ f. & w/ f. & \\
\hline  V1 & & - & -1.729 \\
V2 & \checkmark & - & -0.626 \\
V3 & \checkmark & & 0.179 \\
V4 & & \checkmark & 0.827 \\
V5 & \checkmark & \checkmark & 1.349 \\

\bottomrule
\end{tabular}

}
\caption{B-T scores. Left sub-table: B-T scores of different baselines and our PHDNet. In ``Type'' column, ``OP'' means optimization-based method, while ``FF'' means feed-forward method. Right sub-table: B-T scores of different network structures, in which $G$ (\emph{resp.}, $D$) means generator (\emph{resp.}, discriminator), w/ f. means ``with frequency-related module'', ``-'' means without discriminator.}

\label{tab:bt_score}
\end{table}

\subsection{Ablation Studies} \label{sec:ablation}
We ablate each frequency-related module in our PHDNet, \emph{i.e.}, the ResFFT module in generator $G$ and the frequency branch $D_f$ in discriminator $D$.  We construct different ablated versions according to whether using ResFFT module, whether using discriminator, and whether using frequency branch in the discriminator, leading to in total 5 versions. 
As summarized  in the right sub-table in Table~\ref{tab:bt_score}, we first remove the ResFFT module in the generator and remove the whole discriminator, which is referred to as ``V1". Then, we add ResFFT module in the generator, leading to ``V2". Based on ``V2", we add the discriminator without frequency branch, leading to ``V3". Next, we further add frequency branch to the discriminator, arriving at our full version ``V5". Additionally, based on ``V5", we remove the ResFFT module in the generator and get ``V4". 
Following the way in Section~\ref{sec:user_study}, we conduct user study and employ the B-T model~\cite{bradley1952rank,lai2016comparative} to obtain the overall ranking of all versions.

From the right sub-table in Table~\ref{tab:bt_score}, we can see that the performances without using discriminator (``V1", ``V2") are very poor. Based on ``V2" and ``V3", we can see that even using the simplified discriminator without frequency branch can significantly improve the performance, which demonstrates that it is useful to push the foreground to be indistinguishable from the background. 
The comparison between ``V1" (\emph{resp.}, ``V4")  and ``V2" (\emph{resp.}, ``V5") verifies the effectiveness of the ResFFT module in the generator. The comparison between ``V3" and ``V5" verifies the effectiveness of the frequency branch in the discriminator. 
Together with two frequency-related modules, our full version ``V5" achieves the highest score, which proves that the frequency branch in the discriminator can help the ResFFT module learn to harmonize the frequency feature map.

\begin{figure}[t!]
\centering
\includegraphics[width=0.99\linewidth]{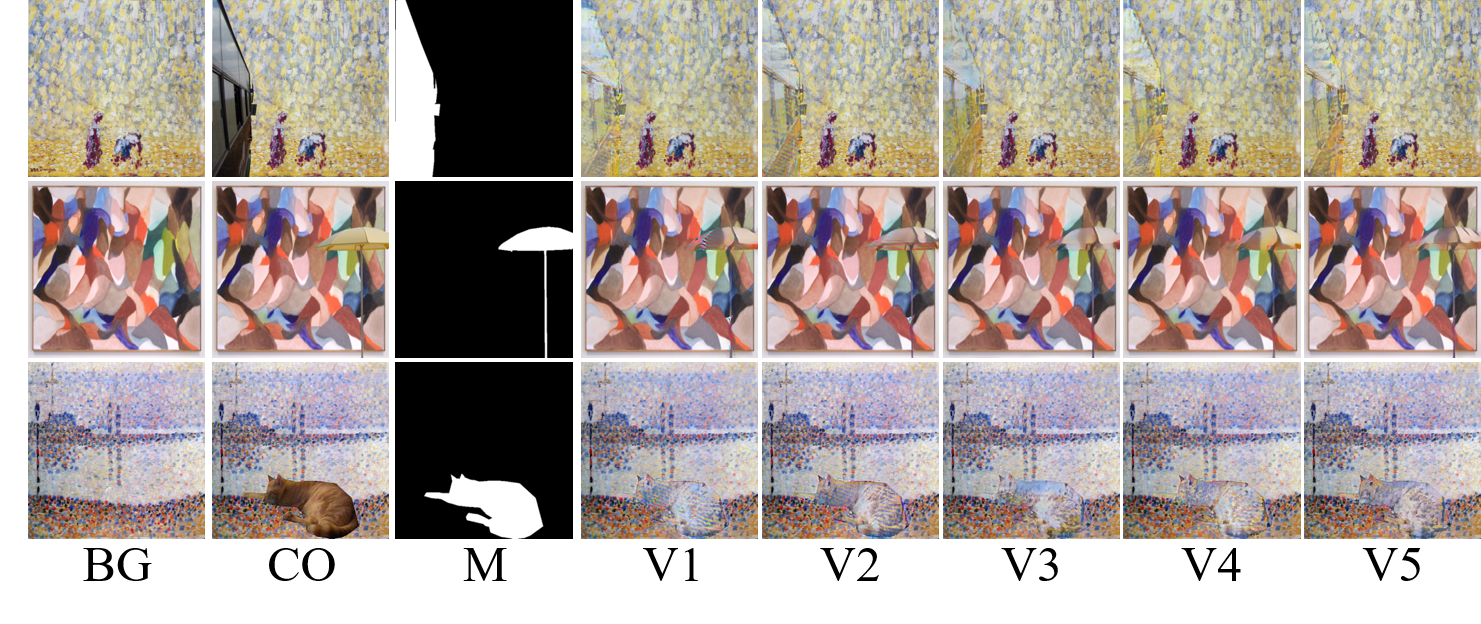}
\caption{Example results of each ablated version.}
\label{fig:ablation}
\end{figure}

In addition, we show the harmonized results of different versions in Figure~\ref{fig:ablation}. One observation is that the generated results without using discriminator (``V1", ``V2") are prone to have artifacts and the discriminator can help enhance the quality of generated images. Another observation is that the frequency-related modules (ResFFT module in the generator and the frequency branch in the discriminator) can collaborate with each other to better transfer the textures/patterns from background image to composite foreground, resulting in more harmonious images. 

\subsection{Hyper-Parameter Analyses}
We investigate the impact of the hyper-parameter in PHDNet, \emph{i.e.}, the patch number $n^2$ in our dual-domain discriminator (see Section~\ref{sec:discriminator}). We provide the visualization results when varying $n$. Details are left to the Supplementary.

\subsection{Visualization of Frequency Maps}
In order to demonstrate the effectiveness of frequency domain learning in our PHDNet intuitively, we visualize the different frequency maps of our frequency-related modules.
The comparison results show that our PHDNet can well transfer the textures from the background style image to the foreground of the composite image, and generate the harmonized image.
Details are also left to the Supplementary.

\subsection{Limitations}
Although our PHDNet can generally produce visually appealing and harmonious results, it may also generate under-stylized results when handling certain types of background styles. 
More discussions and detailed results can be found in the Supplementary.

\section{Conclusion}
In this work, we have introduced frequency domain learning into painterly image harmonization task. We have proposed a novel dual-domain network PHDNet, which contains a dual-domain generator and a dual-domain discriminator. Extensive experiments have demonstrated that our PHDNet has very strong style transfer ability and the stylized foreground is compatible with the background.

\section*{Acknowledgments}
The work was supported by the National Science Foundation of China (62076162), and the Shanghai Municipal Science and Technology Major/Key Project, China (2021SHZDZX0102, 20511100300).

\bibliography{egbib}

\begin{thebibliography}{50}
\providecommand{\natexlab}[1]{#1}

\bibitem[{Bansal, Sheikh, and Ramanan(2018)}]{bansal2017pixelnn}
Bansal, A.; Sheikh, Y.; and Ramanan, D. 2018.
\newblock Pixelnn: Example-based image synthesis.
\newblock \emph{ICLR}.

\bibitem[{Bradley and Terry(1952)}]{bradley1952rank}
Bradley, R.~A.; and Terry, M.~E. 1952.
\newblock Rank analysis of incomplete block designs: I. The method of paired
  comparisons.
\newblock \emph{Biometrika}.

\bibitem[{Cai et~al.(2021)Cai, Zhang, Huang, Geng, Li, and
  Huang}]{cai2021frequency}
Cai, M.; Zhang, H.; Huang, H.; Geng, Q.; Li, Y.; and Huang, G. 2021.
\newblock Frequency domain image translation: More photo-realistic, better
  identity-preserving.
\newblock In \emph{ICCV}.

\bibitem[{Cong et~al.(2021)Cong, Niu, Zhang, Liang, and Zhang}]{bargain}
Cong, W.; Niu, L.; Zhang, J.; Liang, J.; and Zhang, L. 2021.
\newblock {BargainNet}: Background-Guided Domain Translation for Image
  Harmonization.
\newblock In \emph{ICME}.

\bibitem[{Cong et~al.(2022)Cong, Tao, Niu, Liang, Gao, Sun, and
  Zhang}]{Cong_2022_CVPR}
Cong, W.; Tao, X.; Niu, L.; Liang, J.; Gao, X.; Sun, Q.; and Zhang, L. 2022.
\newblock High-Resolution Image Harmonization via Collaborative Dual
  Transformations.
\newblock In \emph{CVPR}.

\bibitem[{Cong et~al.(2020)Cong, Zhang, Niu, Liu, Ling, Li, and
  Zhang}]{cong2020dovenet}
Cong, W.; Zhang, J.; Niu, L.; Liu, L.; Ling, Z.; Li, W.; and Zhang, L. 2020.
\newblock Dovenet: Deep image harmonization via domain verification.
\newblock In \emph{CVPR}.

\bibitem[{Cun and Pun(2020)}]{xiaodong2019improving}
Cun, X.; and Pun, C. 2020.
\newblock Improving the Harmony of the Composite Image by Spatial-Separated
  Attention Module.
\newblock \emph{{IEEE} Transactions on Image Processing}, 29: 4759--4771.

\bibitem[{Deng et~al.(2022)Deng, Tang, Dong, Ma, Pan, Wang, and
  Xu}]{deng2022stytr2}
Deng, Y.; Tang, F.; Dong, W.; Ma, C.; Pan, X.; Wang, L.; and Xu, C. 2022.
\newblock StyTr2: Image Style Transfer with Transformers.
\newblock In \emph{CVPR}.

\bibitem[{Du(2020)}]{du2020much}
Du, L. 2020.
\newblock How much deep learning does neural style transfer really need? an
  ablation study.
\newblock In \emph{WACV}.

\bibitem[{Gatys, Ecker, and Bethge(2016)}]{gatys2016image}
Gatys, L.~A.; Ecker, A.~S.; and Bethge, M. 2016.
\newblock Image style transfer using convolutional neural networks.
\newblock In \emph{CVPR}.

\bibitem[{Goodfellow et~al.(2014)Goodfellow, Pouget-Abadie, Mirza, Xu,
  Warde-Farley, Ozair, Courville, and Bengio}]{goodfellow2014generative}
Goodfellow, I.; Pouget-Abadie, J.; Mirza, M.; Xu, B.; Warde-Farley, D.; Ozair,
  S.; Courville, A.; and Bengio, Y. 2014.
\newblock Generative adversarial nets.
\newblock \emph{NeurIPS}, 27.

\bibitem[{Guo et~al.(2021{\natexlab{a}})Guo, Guo, Zheng, Gu, Zheng, and
  Dong}]{guo2021image}
Guo, Z.; Guo, D.; Zheng, H.; Gu, Z.; Zheng, B.; and Dong, J.
  2021{\natexlab{a}}.
\newblock Image harmonization with transformer.
\newblock In \emph{ICCV}.

\bibitem[{Guo et~al.(2021{\natexlab{b}})Guo, Zheng, Jiang, Gu, and
  Zheng}]{guo2021intrinsic}
Guo, Z.; Zheng, H.; Jiang, Y.; Gu, Z.; and Zheng, B. 2021{\natexlab{b}}.
\newblock Intrinsic Image Harmonization.
\newblock In \emph{CVPR}.

\bibitem[{Hang et~al.(2022)Hang, Xia, Yang, and Liao}]{hang2022scs}
Hang, Y.; Xia, B.; Yang, W.; and Liao, Q. 2022.
\newblock SCS-Co: Self-Consistent Style Contrastive Learning for Image
  Harmonization.
\newblock In \emph{CVPR}.

\bibitem[{Hao, Iizuka, and Fukui(2020)}]{Hao2020bmcv}
Hao, G.; Iizuka, S.; and Fukui, K. 2020.
\newblock Image Harmonization with Attention-based Deep Feature Modulation.
\newblock In \emph{BMVC}.

\bibitem[{He et~al.(2016)He, Zhang, Ren, and Sun}]{he2016deep}
He, K.; Zhang, X.; Ren, S.; and Sun, J. 2016.
\newblock Deep residual learning for image recognition.
\newblock In \emph{CVPR}.

\bibitem[{Huang and Belongie(2017)}]{huang2017arbitrary}
Huang, X.; and Belongie, S. 2017.
\newblock Arbitrary style transfer in real-time with adaptive instance
  normalization.
\newblock In \emph{Proceedings of the IEEE international conference on computer
  vision}.

\bibitem[{Huo et~al.(2021)Huo, Jin, Li, Wu, Lai, Shi, and
  Gao}]{huo2021manifold}
Huo, J.; Jin, S.; Li, W.; Wu, J.; Lai, Y.-K.; Shi, Y.; and Gao, Y. 2021.
\newblock Manifold alignment for semantically aligned style transfer.
\newblock In \emph{ICCV}.

\bibitem[{Kolkin, Salavon, and Shakhnarovich(2019)}]{kolkin2019style}
Kolkin, N.; Salavon, J.; and Shakhnarovich, G. 2019.
\newblock Style transfer by relaxed optimal transport and self-similarity.
\newblock In \emph{CVPR}.

\bibitem[{Lai et~al.(2016)Lai, Huang, Hu, Ahuja, and Yang}]{lai2016comparative}
Lai, W.; Huang, J.; Hu, Z.; Ahuja, N.; and Yang, M. 2016.
\newblock A Comparative Study for Single Image Blind Deblurring.
\newblock In \emph{CVPR}.

\bibitem[{Lalonde and Efros(2007)}]{lalonde2007using}
Lalonde, J.; and Efros, A.~A. 2007.
\newblock Using Color Compatibility for Assessing Image Realism.
\newblock In \emph{ICCV}.

\bibitem[{Li et~al.(2018)Li, Liu, Kautz, and Yang}]{li2018learning}
Li, X.; Liu, S.; Kautz, J.; and Yang, M.-H. 2018.
\newblock Learning linear transformations for fast arbitrary style transfer.
\newblock \emph{arXiv preprint arXiv:1808.04537}.

\bibitem[{Li et~al.(2017{\natexlab{a}})Li, Fang, Yang, Wang, Lu, and
  Yang}]{li2017universal}
Li, Y.; Fang, C.; Yang, J.; Wang, Z.; Lu, X.; and Yang, M.-H.
  2017{\natexlab{a}}.
\newblock Universal style transfer via feature transforms.
\newblock \emph{NeurIPS}.

\bibitem[{Li et~al.(2017{\natexlab{b}})Li, Wang, Liu, and
  Hou}]{li2017demystifying}
Li, Y.; Wang, N.; Liu, J.; and Hou, X. 2017{\natexlab{b}}.
\newblock Demystifying neural style transfer.
\newblock In \emph{IJCAI}.

\bibitem[{Lin et~al.(2014)Lin, Maire, Belongie, Hays, Perona, Ramanan,
  Doll{\'a}r, and Zitnick}]{lin2014microsoft}
Lin, T.-Y.; Maire, M.; Belongie, S.; Hays, J.; Perona, P.; Ramanan, D.;
  Doll{\'a}r, P.; and Zitnick, C.~L. 2014.
\newblock Microsoft coco: Common objects in context.
\newblock 740--755. Springer.

\bibitem[{Ling et~al.(2021)Ling, Xue, Song, Xie, and Gu}]{ling2021region}
Ling, J.; Xue, H.; Song, L.; Xie, R.; and Gu, X. 2021.
\newblock Region-aware adaptive instance normalization for image harmonization.
\newblock In \emph{CVPR}.

\bibitem[{Liu et~al.(2021)Liu, Lin, He, Li, Wang, Li, Sun, Li, and
  Ding}]{liu2021adaattn}
Liu, S.; Lin, T.; He, D.; Li, F.; Wang, M.; Li, X.; Sun, Z.; Li, Q.; and Ding,
  E. 2021.
\newblock Adaattn: Revisit attention mechanism in arbitrary neural style
  transfer.
\newblock In \emph{ICCV}.

\bibitem[{Luan et~al.(2018)Luan, Paris, Shechtman, and Bala}]{luan2018deep}
Luan, F.; Paris, S.; Shechtman, E.; and Bala, K. 2018.
\newblock Deep painterly harmonization.
\newblock In \emph{Computer graphics forum}, 95--106. Wiley Online Library.

\bibitem[{Mahendran and Vedaldi(2015)}]{mahendran2015understanding}
Mahendran, A.; and Vedaldi, A. 2015.
\newblock Understanding deep image representations by inverting them.
\newblock In \emph{CVPR}.

\bibitem[{Mardani et~al.(2020)Mardani, Liu, Dundar, Liu, Tao, and
  Catanzaro}]{mardani2020neural}
Mardani, M.; Liu, G.; Dundar, A.; Liu, S.; Tao, A.; and Catanzaro, B. 2020.
\newblock Neural ffts for universal texture image synthesis.
\newblock \emph{NeurIPS}.

\bibitem[{Park and Lee(2019)}]{park2019arbitrary}
Park, D.~Y.; and Lee, K.~H. 2019.
\newblock Arbitrary style transfer with style-attentional networks.
\newblock In \emph{CVPR}.

\bibitem[{Peng, Wang, and Wang(2019)}]{peng2019element}
Peng, H.-J.; Wang, C.-M.; and Wang, Y.-C.~F. 2019.
\newblock Element-Embedded Style Transfer Networks for Style Harmonization.
\newblock In \emph{BMVC}.

\bibitem[{P{\'{e}}rez, Gangnet, and Blake(2003)}]{poisson}
P{\'{e}}rez, P.; Gangnet, M.; and Blake, A. 2003.
\newblock Poisson image editing.
\newblock \emph{{ACM} Transactions on Graphics}, 22(3): 313--318.

\bibitem[{Ronneberger, Fischer, and Brox(2015)}]{ronneberger2015u}
Ronneberger, O.; Fischer, P.; and Brox, T. 2015.
\newblock U-net: Convolutional networks for biomedical image segmentation.
\newblock In \emph{International Conference on Medical image computing and
  computer-assisted intervention}, 234--241. Springer.

\bibitem[{Roy et~al.(2021)Roy, Chaudhury, Yamasaki, and
  Hashimoto}]{roy2021image}
Roy, H.; Chaudhury, S.; Yamasaki, T.; and Hashimoto, T. 2021.
\newblock Image inpainting using frequency-domain priors.
\newblock \emph{Journal of Electronic Imaging}, 30(2): 023016.

\bibitem[{Shen et~al.(2021)Shen, Yang, Wei, Deng, Huang, Hua, Cheng, and
  Liang}]{shen2021dct}
Shen, X.; Yang, J.; Wei, C.; Deng, B.; Huang, J.; Hua, X.-S.; Cheng, X.; and
  Liang, K. 2021.
\newblock Dct-mask: Discrete cosine transform mask representation for instance
  segmentation.
\newblock In \emph{CVPR}.

\bibitem[{Simonyan and Zisserman(2015)}]{VGG19}
Simonyan, K.; and Zisserman, A. 2015.
\newblock Very deep convolutional networks for large-scale image recognition.
\newblock \emph{ICLR}.

\bibitem[{Sofiiuk, Popenova, and Konushin(2021)}]{sofiiuk2021foreground}
Sofiiuk, K.; Popenova, P.; and Konushin, A. 2021.
\newblock Foreground-aware Semantic Representations for Image Harmonization.
\newblock In \emph{WACV}.

\bibitem[{Song et~al.(2020)Song, Zhong, Qin, and Tu}]{song2020illumination}
Song, S.; Zhong, F.; Qin, X.; and Tu, C. 2020.
\newblock Illumination Harmonization with Gray Mean Scale.
\newblock In \emph{Computer Graphics International Conference}.

\bibitem[{Sunkavalli et~al.(2010)Sunkavalli, Johnson, Matusik, and
  Pfister}]{multi-scale}
Sunkavalli, K.; Johnson, M.~K.; Matusik, W.; and Pfister, H. 2010.
\newblock Multi-scale image harmonization.
\newblock \emph{{ACM} Transactions on Graphics}, 29(4): 125:1--125:10.

\bibitem[{Suvorov et~al.(2022)Suvorov, Logacheva, Mashikhin, Remizova, Ashukha,
  Silvestrov, Kong, Goka, Park, and Lempitsky}]{suvorov2022resolution}
Suvorov, R.; Logacheva, E.; Mashikhin, A.; Remizova, A.; Ashukha, A.;
  Silvestrov, A.; Kong, N.; Goka, H.; Park, K.; and Lempitsky, V. 2022.
\newblock Resolution-robust large mask inpainting with fourier convolutions.
\newblock In \emph{WACV}.

\bibitem[{Tan et~al.(2019)Tan, Chan, Aguirre, and Tanaka}]{artgan2018}
Tan, W.~R.; Chan, C.~S.; Aguirre, H.; and Tanaka, K. 2019.
\newblock Improved ArtGAN for Conditional Synthesis of Natural Image and
  Artwork.
\newblock \emph{IEEE Transactions on Image Processing}, 28(1): 394--409.

\bibitem[{Tsai et~al.(2017)Tsai, Shen, Lin, Sunkavalli, Lu, and
  Yang}]{tsai2017deep}
Tsai, Y.; Shen, X.; Lin, Z.; Sunkavalli, K.; Lu, X.; and Yang, M. 2017.
\newblock Deep Image Harmonization.
\newblock In \emph{CVPR}.

\bibitem[{Wang et~al.(2018)Wang, Girshick, Gupta, and He}]{wang2018non}
Wang, X.; Girshick, R.; Gupta, A.; and He, K. 2018.
\newblock Non-local neural networks.
\newblock In \emph{CVPR}.

\bibitem[{Xu et~al.(2020)Xu, Qin, Sun, Wang, Chen, and Ren}]{xu2020learning}
Xu, K.; Qin, M.; Sun, F.; Wang, Y.; Chen, Y.-K.; and Ren, F. 2020.
\newblock Learning in the frequency domain.
\newblock In \emph{CVPR}.

\bibitem[{Xue et~al.(2012)Xue, Agarwala, Dorsey, and
  Rushmeier}]{xue2012understanding}
Xue, S.; Agarwala, A.; Dorsey, J.; and Rushmeier, H.~E. 2012.
\newblock Understanding and improving the realism of image composites.
\newblock \emph{{ACM} Transactions on Graphics}, 31(4): 84:1--84:10.

\bibitem[{Yang and Soatto(2020)}]{yang2020fda}
Yang, Y.; and Soatto, S. 2020.
\newblock Fda: Fourier domain adaptation for semantic segmentation.
\newblock In \emph{CVPR}.

\bibitem[{Yu et~al.(2021)Yu, Zhan, Lu, Pan, Ma, Xie, and Miao}]{yu2021wavefill}
Yu, Y.; Zhan, F.; Lu, S.; Pan, J.; Ma, F.; Xie, X.; and Miao, C. 2021.
\newblock Wavefill: A wavelet-based generation network for image inpainting.
\newblock In \emph{ICCV}.

\bibitem[{Zhang, Wen, and Shi(2020)}]{zhang2020deep}
Zhang, L.; Wen, T.; and Shi, J. 2020.
\newblock Deep image blending.
\newblock In \emph{WACV}.

\bibitem[{Zhu et~al.(2015)Zhu, Kr{\"{a}}henb{\"{u}}hl, Shechtman, and
  Efros}]{zhu2015learning}
Zhu, J.; Kr{\"{a}}henb{\"{u}}hl, P.; Shechtman, E.; and Efros, A.~A. 2015.
\newblock Learning a Discriminative Model for the Perception of Realism in
  Composite Images.
\newblock In \emph{ICCV}.

\end{thebibliography}


\begin{thebibliography}{6}
\providecommand{\natexlab}[1]{#1}

\bibitem[{Deng et~al.(2022)Deng, Tang, Dong, Ma, Pan, Wang, and
  Xu}]{deng2022stytr2}
Deng, Y.; Tang, F.; Dong, W.; Ma, C.; Pan, X.; Wang, L.; and Xu, C. 2022.
\newblock StyTr2: Image Style Transfer with Transformers.
\newblock In \emph{CVPR}.

\bibitem[{Lin et~al.(2014)Lin, Maire, Belongie, Hays, Perona, Ramanan,
  Doll{\'a}r, and Zitnick}]{lin2014microsoft}
Lin, T.-Y.; Maire, M.; Belongie, S.; Hays, J.; Perona, P.; Ramanan, D.;
  Doll{\'a}r, P.; and Zitnick, C.~L. 2014.
\newblock Microsoft coco: Common objects in context.
\newblock 740--755. Springer.

\bibitem[{Liu et~al.(2021)Liu, Lin, He, Li, Wang, Li, Sun, Li, and
  Ding}]{liu2021adaattn}
Liu, S.; Lin, T.; He, D.; Li, F.; Wang, M.; Li, X.; Sun, Z.; Li, Q.; and Ding,
  E. 2021.
\newblock Adaattn: Revisit attention mechanism in arbitrary neural style
  transfer.
\newblock In \emph{ICCV}.

\bibitem[{Luan et~al.(2018)Luan, Paris, Shechtman, and Bala}]{luan2018deep}
Luan, F.; Paris, S.; Shechtman, E.; and Bala, K. 2018.
\newblock Deep painterly harmonization.
\newblock In \emph{Computer graphics forum}, 95--106. Wiley Online Library.

\bibitem[{Park and Lee(2019)}]{park2019arbitrary}
Park, D.~Y.; and Lee, K.~H. 2019.
\newblock Arbitrary style transfer with style-attentional networks.
\newblock In \emph{CVPR}.

\bibitem[{Tan et~al.(2019)Tan, Chan, Aguirre, and Tanaka}]{artgan2018}
Tan, W.~R.; Chan, C.~S.; Aguirre, H.; and Tanaka, K. 2019.
\newblock Improved ArtGAN for Conditional Synthesis of Natural Image and
  Artwork.
\newblock \emph{IEEE Transactions on Image Processing}, 28(1): 394--409.

\end{thebibliography}

\end{document}


\maketitle

In the supplementary, we will first introduce the implementation details in Section~\ref{sec:implementation}.
Next, we will  analyze the impact of the hyper-parameter in Section~\ref{sec:hyper}. 
We will visualize the frequency feature maps extracted by our frequency-related modules in Section~\ref{sec:freq_feat}. 
Then, we will provide more harmonized results compared to the strong baselines in Section~\ref{sec:baseline}. 
Finally, we will discuss the limitations of our PHDNet in Section~\ref{sec:limitation}.

\section{Implementation Details} \label{sec:implementation}
We conduct experiments on COCO \cite{lin2014microsoft} and WikiArt \cite{artgan2018}.
COCO is a benchmark dataset with instance segmentation annotation for 80 object categories, while WikiArt is a large-scale digital art dataset with 27 different styles.
Therefore we utilize these two datasets to produce the composite images, in which the photographic foreground objects are from COCO while the painterly backgrounds are from WikiArt.
Specifically, we randomly select an object in one image from COCO with foreground ratio in range $[0.05, 0.3]$. Then we cut it out using the instance annotation as the foreground object, and paste it onto a randomly selected background image from WikiArt, leading to an inharmonious composite image. 
Following the settings in \cite{artgan2018}, we have 57,025 background images for training and 24,421 for testing.

The architecture of our dual-domain generator $G$ is clearly described in Section 3.1 in the main paper. Our dual-domain discriminator $D$ is built upon downsample (DS) blocks and upsample (US) blocks. Specifically, we apply six DS blocks inside $D_s$, in which each block contains a convolution (Conv) layer with kernel size of four and a stride of two followed by a batch normalization (BN) layer and a LeakyReLU activation. For the frequency branch $D_f$, we apply three DS blocks, in which the structure of each block is the same as $D_s$. At the end of $D_f$, we insert a fully connected layer to obtain the frequency feature vector of each patch. $D_a$ is a small-scale auto-encoder with two DS blocks and two US blocks. Each DS block of $D_a$ contains a Conv layer with kernel size of three and a stride of one, a BN layer, and a LeakyReLU activation sequentially. 
While each US block has the same structure as the DS block of $D_a$ except the ReLU activation. 
We set the patch number in our dual-domain discriminator as $n=4$.


Our network is implemented using Pytorch 1.7.0 and trained using Adam optimizer with learning rate of $2e{-4}$ on ubuntu 18.04 LTS operation system, with 32GB memory, Intel Core i7-9700K CPU, and two GeForce GTX 2080 Ti GPUs. We resize the input images as $256 \times 256$ during the training phase. 
As our network is fully convolutional, it can be applied to images of any size in the test phase.

\section{Hyper-parameter Analysis} \label{sec:hyper}
We investigate the impact of the hyper-parameter in our network, \emph{i.e.}, the patch number $n^2$ in our dual-domain discriminator (Section 3.2 in the main paper).

\begin{figure}[t!]
\centering
\includegraphics[width=0.99\linewidth]{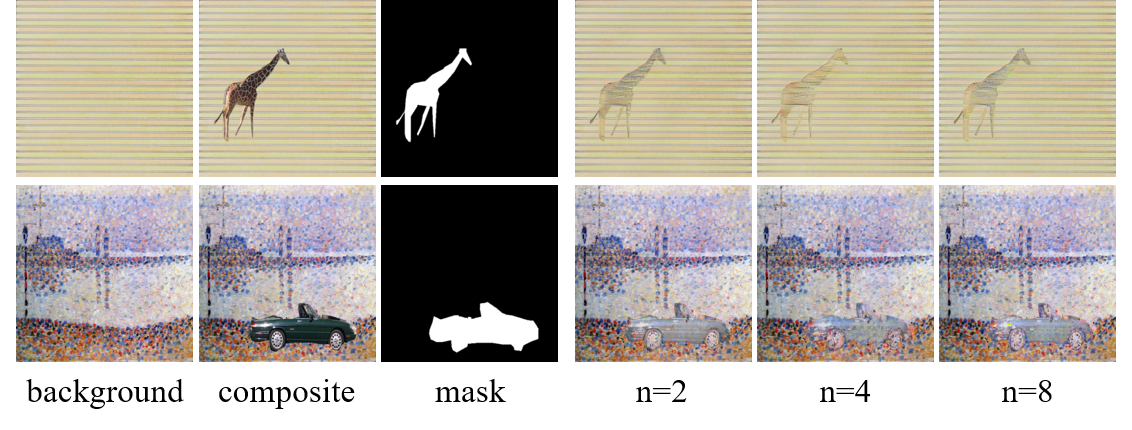}
\caption{Example results of different patch numbers $n^2$ in our $D_f$.}
\label{fig:ablation_n}
\end{figure}

The impact of using different $n$ is shown in Figure~\ref{fig:ablation_n}. When $n= 2,4,8$, the number of patches is $2 \times 2$, $4 \times 4$, $8 \times 8$ respectively, corresponding to the inharmonious region mask with size $2 \times 2$, $4 \times 4$, $8 \times 8$ respectively. When the number of patches is large ($8 \times 8$), patch size is very small and each patch does not contain adequate information. When the number of patches is small ($2 \times 2$), the foreground patches may include much background information and cannot precisely enclose the foreground object, which makes it less effective to distinguish foreground and background patches. Therefore, $n=4$ is a reasonable choice. 
From Figure~\ref{fig:ablation_n}, we can also observe that the results obtained using $n=4$ are better than those obtained using $n=2$ or $n=8$, which complies with our above analyses. 

\begin{figure}[t!]
\centering
\includegraphics[width=0.99\linewidth]{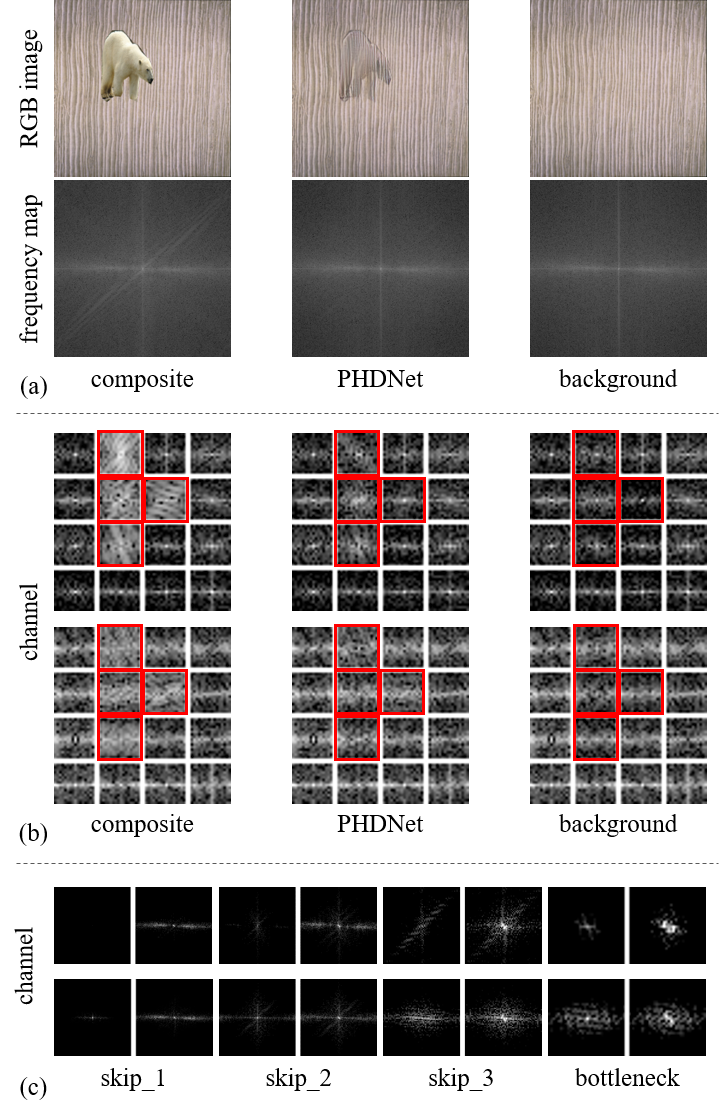}
\caption{Visualization of frequency maps. We show (a) the RGB images (\emph{resp.}, frequency maps) of composite image, our harmonized image, and background image in the top (\emph{resp.}, bottom) row, (b) $4 \times 4$ frequency feature maps of $4 \times 4$ patches of composite image, our harmonized image, and background image, in which the foreground patches are outlined in red, (c) frequency feature maps before (\emph{resp.}, after) our ResFFT module in the left (\emph{resp.}, right) subfigure in each encoder layer (3 skip connections and 1 bottleneck). }
\label{fig:frequency}
\end{figure}

\section{Visualization of Frequency Maps} \label{sec:freq_feat}
In this section, we demonstrate the effectiveness of frequency domain learning in our network intuitively.
First, we show an example triplet of composite image, our harmonized image, and background image in Figure~\ref{fig:frequency}(a). We apply \emph{FFT} to these images and visualize the obtained FFT magnitude in log scale, which is referred to as frequency map for ease of description. 
The frequency maps of three images are shown in Figure~\ref{fig:frequency}(a). Since the example background image has very regular textures, its frequency map exhibits bright lines clearly. In the composite image, as the inserted object has considerably different textures from the background, its frequency map changes a lot, compared with that of background image. After harmonization, the foreground is filled with background texture and naturally blended in the background, so the frequency map is restored and close to that of background image.

In Figure~\ref{fig:frequency}(b), we visualize the frequency feature map $F_{df}^{i,j}$ of each patch in the discriminator, which represents the frequency information of the $(i,j)$-th patch. As FFT is applied to each channel in $F_{dm}^{i,j}$ independently to get $F_{df}^{i,j}$, we can visualize the frequency feature map for each channel. Here, we only visualize two channels and the observations on the other channels are similar. 
Recall that we set $n=4$ by default, so an image is divided into $4\times 4$ patches. We show $16$ frequency feature maps for $16$ patches in the composite image, our harmonized image, and background image respectively. 
For the composite image and our harmonized image, the frequency feature maps of background patches are similar to those in the background image. 
For the composite image, the frequency feature maps of foreground patches are far from those of background patches, which means that the foreground lacks the texture/pattern in the background. Thus, the discriminator can easily distinguish foreground patches from background patches on the premise of frequency information. For our harmonized image, the frequency feature maps of foreground patches are more consistent with those of background patches. This demonstrates that our generator has the ability to fool the discriminator by generating harmonized image with consistent foreground and background frequency information. 


In Figure~\ref{fig:frequency}(c), to investigate the harmonization effect of our ResFFT module, we visualize the frequency feature map $F_{gf}^l$ before using our ResFFT module and the frequency feature map $\hat{F}_{gf}^l$ after using our ResFFT module in the generator, in which $l$ means the $l$-th encoder layer. Similar to Figure~\ref{fig:frequency}(b), we only visualize two channels and the observations on the other channels are similar. 
We show the visualization results for all four encoder layers (3 skip connections and 1 bottleneck).
As illustrated in Figure~\ref{fig:frequency}(c), after going through the ResFFT module, some new bright lines appear in the frequency feature maps, or some bright regions become brighter and cleaner. These visualization results demonstrate that our ResFFT module can add new textures or strengthen the existing textures by manipulating the frequency feature maps, so that the foreground in the harmonized image has more compatible textures with the background. 




\begin{figure*}
\centering
\includegraphics[width=0.99\linewidth]{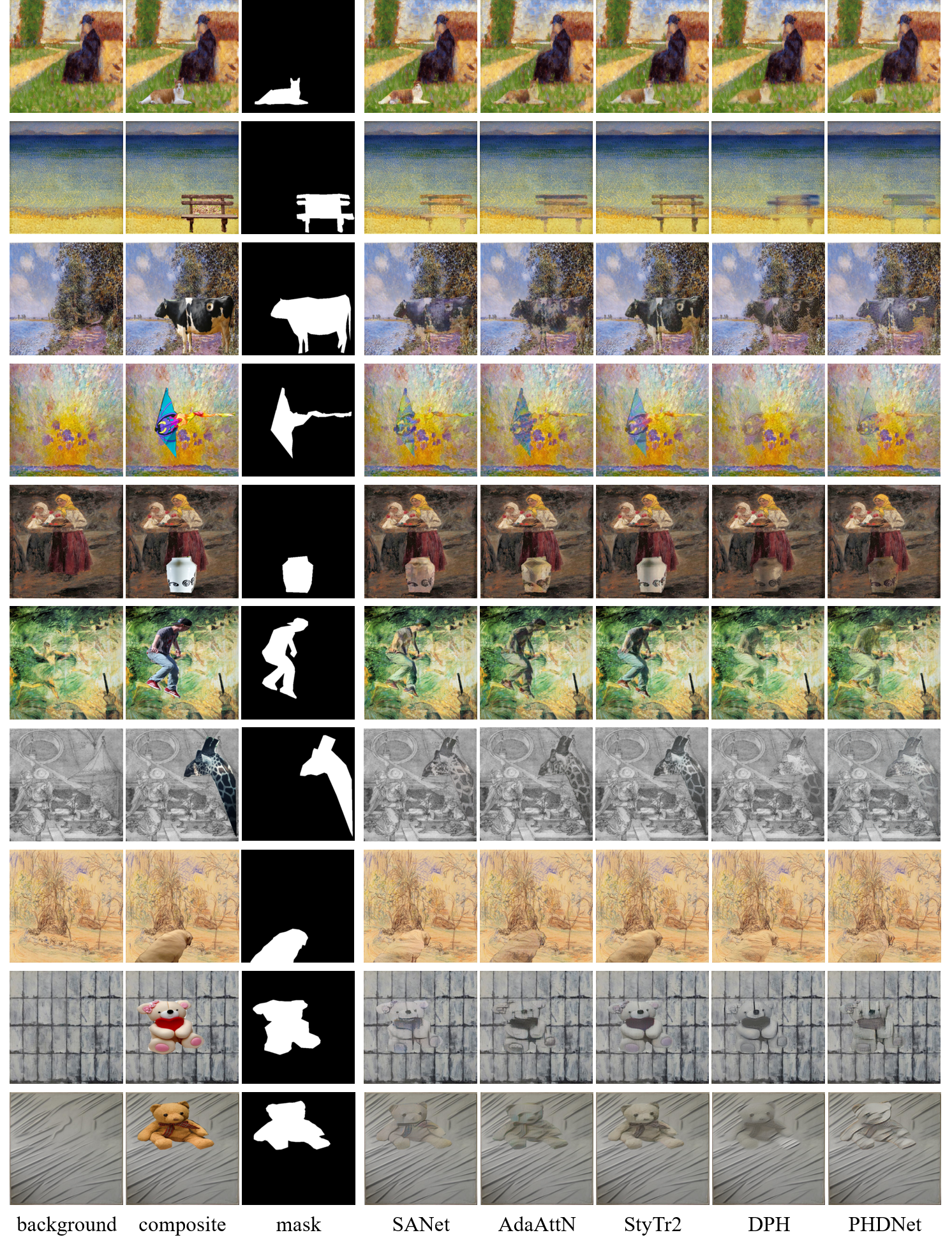}
\caption{From left to right are the background image, composite image, composite foreground mask, the harmonized results of SANet~\cite{park2019arbitrary}, AdaAttN~\cite{liu2021adaattn}, StyTr2~\cite{deng2022stytr2}, DPH~\cite{luan2018deep} and our PHDNet.}
\label{fig:baseline}
\end{figure*}

\begin{figure}[tb]
\centering
\includegraphics[width=0.99\linewidth]{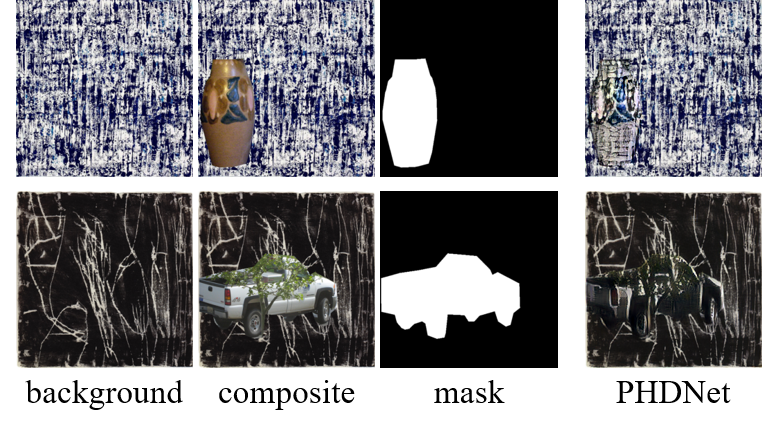}
\caption{Example failure cases of our PHDNet.}
\label{fig:failure}
\end{figure}

\section{More Comparison with Baselines} \label{sec:baseline}
We select the strong baselines SANet~\cite{park2019arbitrary}, AdaAttN~\cite{liu2021adaattn},
StyTr2~\cite{deng2022stytr2}, and DPH~\cite{luan2018deep} from two groups of baselines, in which DPH is from the painterly image harmonization group while the rest are from the artistic style transfer group. 
In Figure~\ref{fig:baseline}, we show the harmonized results generated by baseline methods and our PHDNet.
Compared with these strong baselines, our PHDNet can generally transfer the style from background to foreground better, producing more harmonious and visually appealing results. For example, our PHDNet can transfer fine-grained style from background while maintaining the content structure of foreground object (\emph{e.g.}, row 1, 2, 3, 4, 5), while the baseline method may fail to stylize the foreground object or blur the content structure. Moreover, our PHDNet is able to adjust the foreground color to be more compatible with background (\emph{e.g.}, row 6, 7) than the baseline methods. our PHDNet is also capable of making subtle changes to the foreground object to fit the background style. For instance, in row 9, the shapes of bear body parts (\emph{e.g.}, nose, heart, and paws) are converted to square and several vertical lines are added to the bear face, due to the square bricks in the background.
In some challenging cases (\emph{e.g.}, row 10) where baselines produce very poor results, our PHDNet can still generate satisfactory results. Overall, in our harmonized images, the foreground is naturally blended in the background and it is hard to identify which object is the inserted object.

\section{Limitations} \label{sec:limitation}
We show the limitations of our PHDNet in Figure~\ref{fig:failure}. For the background image with monotonous and highly contrastive color, our PHDNet could not adjust the foreground style to be completely compatible with the background style. For example, the vase (row 1) and the car (row 2) still have original colors after harmonization. We will explore this problem in the future.

\bibliography{egbib}